\documentclass[11pt,a4paper]{article}
\usepackage{authblk}
\usepackage[hyperref]{ranlp2021}
\usepackage{times}
\usepackage{latexsym}

\usepackage{todonotes}
\usepackage{amsfonts}
\usepackage{amsmath}
\usepackage{multirow}
\usepackage{adjustbox}
\usepackage{booktabs}
\usepackage{arydshln}
\usepackage{titlesec}
\usepackage{soul}
\usepackage{subfig}

\titlespacing{\paragraph}{%
  0pt}{%              left margin
  0.05\baselineskip}{% space before (vertical)
  1em}%               space after (horizontal)

\usepackage{microtype}

\aclfinalcopy % Uncomment this line for the final submission
\def\aclpaperid{158} %  Enter the acl Paper ID here

\makeatletter
\newcommand\footnoteref[1]{\protected@xdef\@thefnmark{\ref{#1}}\@footnotemark}
\makeatother

\renewcommand{\vec}[1]{\ensuremath{\mathbf{#1}}}

\DeclareMathOperator*{\argmax}{argmax~}
\title{Are the Multilingual Models Better? \\ Improving Czech Sentiment with Transformers}

	\author[1,2]{\bf Pavel P\v{r}ib\'{a}\v{n}}
	\author[1,2]{\bf  Josef Steinberger}
	\affil[ ]{University of West Bohemia, Faculty of Applied Sciences, Czech Republic}
	\affil[1]{NTIS -- New Technologies for the Information Society,}
	\affil[2]{Department of Computer Science and Engineering,}
	\affil[  ]{\tt	\{pribanp,jstein\}@kiv.zcu.cz}
	\affil[  ]{\tt http://nlp.kiv.zcu.cz}

\date{}

\begin{document}
\maketitle
\begin{abstract}
In this paper, we aim at improving Czech sentiment with transformer-based models and their multilingual versions. More concretely, we study the task of polarity detection for the Czech language on three sentiment polarity datasets. We fine-tune and perform experiments with five multilingual and three monolingual models.  We compare the monolingual and multilingual models' performance, including comparison with the older approach based on recurrent neural networks. Furthermore, we test the multilingual models and their ability to transfer knowledge from English to Czech (and vice versa) with zero-shot cross-lingual classification. Our experiments show that the huge multilingual models can overcome the performance of the monolingual models. They are also able to detect polarity in another language without any training data, with performance not worse than 4.4 \% compared to state-of-the-art monolingual trained models.  Moreover, we achieved new state-of-the-art results on all three datasets.
\end{abstract}
% -------
% Done
% In this paper we aim at improving Czech sentiment analysis, more concretely the task of polarity detection. 
% In this paper we compare different models for polarity detection of Czech text using three Czech datasets. 

\section{Introduction}

In recent years, BERT-like models \cite{devlin-etal-2019-bert} based on the Transformer architecture \cite{attention-all-transformer} and generalized language models brought a significant improvement in performance in almost any NLP task \cite{raffel2019exploring}, especially in English. Despite this fact, not much work has been recently done in sentiment analysis for the Czech language with the latest Transformer models. We partly fill this gap by focusing on the \textit{Sentiment Classification} task, also known as \textit{Polarity Detection}.

\par Polarity detection is a classification task where the goal is to assign a sentiment polarity to a given text. The \textit{positive, negative} and \textit{neutral} classes are usually used as the polarity labels. The polarity can also be defined with a different number of labels, i.e., fine-grained sentiment analysis \cite{liu2012sentiment}.

\par The models based on BERT were almost exclusively trained for English, limiting their usage to other languages. Recently, however, their cross-lingual adaptions like mBERT \cite{devlin-etal-2019-bert}, mT5 \cite{xue2020-mt5}, XLM \cite{lample2019cross-xlm} or XLM-R \cite{xlm-r} emerged along with other non-English monolingual versions, for example, Czech \cite{sido2021czert}, French \cite{martin2019camembert, le2019flaubert}, Arabic \cite{safaya2020kuisail-arabicbert}, Romanian \cite{dumitrescu2020birth-romanianBERT}, Dutch \cite{dutch-BERT} or Finnish \cite{finish-bert-2019}.

\par \textcolor{black}{Our motivation is to reveal the performance limits of the current SotA transformer-based models on the Czech polarity detection task, check the ability of the multilingual models to transfer knowledge between languages and unify the procedure and data that enable the correct future evaluation of this task.}
% \st{Our goal is to fine-tune the currently available transformer-based models, reveal and compare performance differences between the monolingual and multilingual models and check their ability to transfer knowledge between languages with the zero-shot cross-lingual classification.}}
\par In this paper, we focus on the task of polarity detection applied on Czech text by comparing the performance of seven pre-trained transformer-based models (both monolingual and multilingual) on three Czech datasets.
% \st{Our goal is to improve the results of the polarity detection task for Czech and give an insight into the currently available models their performance and settings}.
We fine-tune each model on 
%every of the three datasets (for three and two classes) 
each dataset
and we provide a comprehensive survey of their performance.
% Recently, for Czech polarity there was not much workIn recent years, there was not much work done with transfomers on focused on Czech sentiment, this work is first of its type for Czech.
Our experiments show the effectiveness of the Transformer models that significantly outperform the older approaches based on recurrent neural networks. We observe that the monolingual models can be notably outperformed by the multilingual models, but only by those with
%remarkably 
much more parameters. Moreover, we achieve new state-of-the-art results on all three evaluated datasets. 

We are also interested in the ability of the multilingual models to transfer knowledge between languages and its usability for polarity detection. Thus, we perform zero-shot cross-lingual classification, fine-tune four cross-lingual transformer-based models on the English dataset and then test the models on Czech data. We also perform the same experiment in the reverse direction, i.e., from Czech to English. The results reveal that the \texttt{XLM-R-Large} model (fine-tuned solely on English) can achieve very competitive results that are only about 4 \% worse than the SotA model fine-tuned by us on Czech data. To the best of our knowledge, this is the first paper that performs zero-shot cross-lingual polarity detection for the Czech language.

We also noticed that the comparison with the previous works is rather problematic and thus, we provide a split for all Czech datasets that allows comparing future works much easier. Our code and pre-trained models are publicly available\footnote{\label{note:github}\url{https://github.com/pauli31/improving-czech-sentiment-transformers}}.

\par \textcolor{black}{Our main contributions are the following: 1) We provide the comprehensive performance comparison of the currently available transformer-based models for the Czech language on the polarity detection task along with the models' optimal settings. 2) We test the ability of the multilingual models to transfer knowledge between Czech and English.  3) We release all the fine-tuned models and code freely for research purposes and we provide a data split that allows future comparison and evaluation. Furthermore, we achieved new state-of-the-art results for all three evaluated datasets.}

% SOTa

% ==============Done
% 1 tady rict ze existujou multilingual ze zacali vznikat s nima le ze je tu i velke mnozstvi monolingual modelu 
% todo nlp progress citace
% Our code and pre-trained models are publicly avilable \footnote{github ANONYMOUS SUBMISSION} mozna to dat do abstracktu
% \par The previous approaches   pouzivali transfer learning nejdriv nekotnextualizovane word2vec, fasttext, pak ELMO ULMFit BERT atd. citace, vsechny a vsechny byli uspesne aplikovane na klasifikacni problemy resp. detekci polarity ocitovat je
% citovat attention is all you need nekde u tech transfomeru
% todo citace z finetuinng sentiemntu

% citovat mt5 https://arxiv.org/abs/2010.11934 precist si tam related work a zminit z toho ze pak se zacali delat ty multilingual modely jako mBERT, mBART, SlavicBERT, mt5 apod

% nestihl sem
% Take porovnavame prinos nekterych triku z jinych clanku pro detekci sentimentu a to zda fungujou nebo ne

\section{Related Work}
\par The previous approaches \cite{kim-2014-convolutional, oh-lstm, cliche-2017-bb, baziotis-etal-2017-datastories-semeval,gray2017gpu,conneau-etal-2017-deep} for English polarity detection and other related tasks mostly relied on transfer learning and pre-trained word embeddings such as word2vec \cite{mikolov2013w2v} and fastText \cite{bojanowski-etal-2017-enriching-fasttext} in combinations with Convolutional Neural Networks (CNN) or Long Short-Term Memory (LSTM) \cite{lstm-1997}, eventually in conjunction with the modified attention mechanism \cite{bahdanau-attention, rocktaschel2015reasoning, raffel2015feed}. Furthermore, the new contextualized word representations such as CoVe \cite{nips-cove} or ELMo \cite{peters-etal-2018-deep} and pre-trained language model ULMFiT \cite{howard-ruder-2018-universal} were successfully applied to the polarity detection. Finally, the latest transformer-based models like BERT  \cite{devlin-etal-2019-bert}, GPT \cite{radford2018improving}, RoBERTa \cite{liu2019roberta} or T5 \cite{2020-t5} that are all in general trained on language modeling tasks proved their performance superiority for English over all previous approaches, 
%as is shown, 
for example
in \cite{sun2019fine}. These models are pre-trained on  \textcolor{black}{ a modified language modeling tasks with a huge amount of unlabeled data. In the end, they are fine-tuned for a specific downstream task.}

The initial works on Czech polarity detection and sentiment analysis usually used lexical features \cite{steinberger-etal-2011-creating-trian, veselovska2012creating} or Bag-of-Words text representations along with the Naive Bayes or logistic regression classifiers \cite{habernal-etal-2013-sentiment} or a combination of supervised and unsupervised approach \cite{brychcin-habernal-2013-unsupervised-sentiment}. \citet{SLON-Lenc2016Neural} applied CNN using the architecture from \cite{kim-2014-convolutional} and the LSTM neural network to all three datasets that we use in this paper.
% Despite the use of neural network they did not outperform the mentioned unsupervised approach that used more basic classifiers.
Another usage of LSTM neural network with the self-attention mechanism \cite{humphreys2016attentional}
% for the CSFD dataset
can be found in \cite{libovicky2018solving}. Similarly, \citet{sido-curic} tried to use curriculum learning with 
%the 
CNN and LSTM. 
%neural network

% on the CSFD dataset.
%The authors in 
\citet{kybernetika-bert} pre-trained a BERT-based model for polarity detection
% on a boosted dataset from the CSFD\footnote{This is the same webpage TODO ze je stejna jako pro CSFD ale uplne jina data}
with an improved pooling layer and distillation of knowledge technique. The most recent application of the Transformer model is in \cite{sido2021czert}. The authors created a BERT model for Czech and, as one of the evaluation tasks, they performed polarity detection on the FB and CSFD datasets.
% The more detailed survey of older approaches for Czech sentiment analysis can be found in \cite{cano2019sentiment}.

\par \textcolor{black}{To the best of our knowledge, there are no previous works that focus on the zero-shot cross-lingual polarity detection task in the Czech language. The recent related work can be found in \cite{eriguchi2018zero}, where the authors use the neural machine translation encoder-based model and English data to perform zero-shot cross-lingual sentiment classification on French. In \cite{eriguchi2018zero} the authors performed the zero-shot classification from Slovene to Croatian. Another related work can be found in \cite{wang-banko-2021-practical,DBLP:journals/corr/abs-2006-06402}.}

% podivat se po tom sentiment datsetu na cross-lingual

% ================Zero-shot Cross-lingual papers

% ================= DONE
% https://link.springer.com/chapter/10.1007/978-3-030-26061-3_45
% https://link.springer.com/chapter/10.1007/978-3-030-59430-5_5
% https://ufal.ms.mff.cuni.cz/~rosa/2018/docs/ITAT_2018_paper_14_1.pdf
% https://arxiv.org/abs/1901.02780
% https://link.springer.com/chapter/10.1007/978-3-030-59430-5_5 tohle precist
% https://dspace.cvut.cz/handle/10467/83127 tohle je bakalarka
% do releted work pak napisu primo ty papery ktere se venovali sentimentu na cestine
% TODO nas Czert citovat

% zacnu ze sentiment je stara uloha citace pang a lee blabla, v cestine ho resily Habi(citace) ktery udelali dataset a initial experimenty s BoW words, pak reknu sidaka 
% pak nakonec zminim ty vocasy z kyberny
% ================= TODO

% TODO poggolgit jestli nekdo porovnaval transformery na sentimentu
% TODO mozna citovat alexandru
% TODO mozna citovat aspect based pepa
% https://github.com/TheophileBlard/french-sentiment-analysis-with-bert delal to pro francouzstinu ale neni na to paper asi
% todo co mt5 kouknout na related work
% v related by take melo byt pokud najdu nejaka aplikace cross-lingual modelu na jine jazyky v sentimentu

% ============== Poznamky
% Related work clanek z kyberny
% stahli hodne dat z CSFD pro fine utning, pak meli hodne nacrawlovanejch data
% natrenovali BERTA na cestine na velkejch dataech, udelali distilaci s tema stazenejma datama a pridali tam 
% pooling vrstvu ktera funguje asi lip
% vyhodnotili to na binarnich test datech od habiho

\section{Data}
\label{sec:data}
To the best of our knowledge, there are three Czech publicly available datasets for the polarity detection task: (1) movie review dataset (CSFD), (2) Facebook dataset (FB) and (3) product review dataset (Mallcz), all of them come from \cite{habernal-etal-2013-sentiment} and each text sample is annotated with one of three\footnote{The FB dataset also contains 248 samples with a fourth class called \textit{bipolar}, but we ignore this one.} labels, i.e., \textit{positive, neutral} and \textit{negative}, see Table \ref{tab:dataset-distribution} for the class distribution. For the cross-lingual experiments we use the two-class English movie review dataset (IMDB) \cite{maas-etal-2011-learning-imdb}.

\begin{table}[ht!]
\catcode`\-=12
\begin{adjustbox}{width=0.9\linewidth,center}
\begin{tabular}{crrrrr}
\toprule
Dataset                 & Part  & \multicolumn{1}{c}{Positive}         & \multicolumn{1}{c}{Negative}       & \multicolumn{1}{c}{Neutral}                        & \multicolumn{1}{c}{Total} \\ \midrule
\multirow{4}{*}{CSFD}   & train & 22 117   & 21 441  & 22 235                & 65 793                     \\
                        & dev   & 2 456   & 2 399  & 2 456                  & 7 311                      \\
                        & test  & 6 324    & 5 876  & 6 077                & 18 277                     \\ \cdashline{2-6}
                        & total & 30 897   & 29 716  & 30 768               & 91 381                     \\ \midrule
\multirow{4}{*}{FB}     & train & 1 605    & 1 227  & 3 311                 & 6 143                      \\
                        & dev   & 171    & 151    & 361                  & 683                       \\
                        & test  & 811     & 613    & 1 502                & 2 926                      \\ \cdashline{2-6}   
                        & total & 2 587    & 1 991   & 5 174                  & 9 752                      \\ \midrule
\multirow{4}{*}{Mallcz} & train & 74 100  & 7 498    & 23 022                & 104 620                    \\
                        & dev   & 8 253   & 848     & 2 524                  & 11 625                     \\
                        & test  & 20 624   & 2 041    & 6 397                 & 29 062                     \\ \cdashline{2-6}
                        & total & 102 977  & 10 387   & 31 943                & 145 307                    \\ \midrule
\multirow{3}{*}{IMDB}   & train & 12 500   & 12 500  & \multicolumn{1}{c}{-} & 25 000                     \\
                        & test  & 12 500   & 12 500  & \multicolumn{1}{c}{-} & 25 000                     \\ \cdashline{2-6}
                        & total & 25 000   & 25 000  & \multicolumn{1}{c}{-} & 50 000        \\ \bottomrule            
\end{tabular}
\end{adjustbox}
\caption{Datasets statistics.}
\label{tab:dataset-distribution}
\end{table}

% TODO pokud ten paper bude moc dlouhej tak tady muzu ubrat ten popis a nerikat podle ceho je to anotovane
\par The \textbf{FB} dataset contains 10k random posts from nine different Facebook pages that were manually annotated by two annotators.  The \textbf{CSFD} dataset is created from 90k Czech movie reviews from the Czech movie database\footnote{\url{https://www.csfd.cz}} that were downloaded and annotated according to their star rating (0–2 stars as
\textit{negative}, 3–4 stars as \textit{neutral}, 5–6 stars as \textit{positive}). The \textbf{Mallcz} dataset consists of 145k users' reviews of products from Czech e-shop\footnote{\url{https://www.mall.cz}}, the labels are assigned according to the review star rating on the scale 0-5, where the reviews with 0-3 stars are labeled as \textit{negative}, 4 stars as \textit{neutral} and 5 stars as \textit{positive}. The English \textbf{IMDB} dataset includes 50k movie reviews scraped from the Internet Movie Database\footnote{\url{https://www.imdb.com}} with \textit{positive} and \textit{negative} classes split into training and testing parts of equal size.

\par Since there is no official partitioning for the Czech datasets, we split them into training, development and testing parts with the same class distribution for each part as it is in the original dataset, see Table \ref{tab:dataset-distribution}. For the Mallcz and CSFD datasets, we use the following ratio: 80 \% for training data, 20 \% for testing data, for the FB dataset, it is 70 \% and 30 \%, respectively and 10 \% from the training data (for all datasets) is used as the development data. We used different split ratio for the FB dataset because it is approximately ten and sixteen times smaller than the CSFD and Mallcz datasets, respectively and we did not want to reduce the size of the testing data too much.

\section{Models Description}
We performed exhaustive experiments with transformed-based models and in order to compare them with the previous works, we also implemented the older models (baseline models) that include the logistic regression classifier and the BiLSTM neural network.

\subsection{Baseline Models}
We re-implemented the best models from \cite{habernal-etal-2013-sentiment}, i.e., logistic regression classifier (\textbf{lrc}) with character n-grams (in a range from 3-grams to 6-grams), word uni-grams and bi-grams features. The second baseline model is the \textbf{LSTM} model partly inspired by \cite{baziotis-etal-2017-datastories-semeval}. Its input is a sequence of $t$ tokens represented as a matrix $\vec{M} \in \mathbb{R}^{t \times d}$, where $d=300$ is a dimension of the Czech pre-trained fastText word embeddings \cite{bojanowski-etal-2017-enriching-fasttext}\footnote{Available at \url{https://fasttext.cc/docs/en/crawl-vectors.html}}. The maximal size of the input vocabulary is set to 300 000. The input is passed into the trainable embedding layer that is followed by two BiLSTM \cite{bi-LSTM} layers and after each, the dropout \cite{srivastava2014dropout} is applied. After the two BiLSTM layers, the self-attention mechanism is applied. The output is then passed to a fully-connected softmax layer. An output of the softmax layer is a probability distribution over the possible classes. We use the Adam \cite{Kingma-adam} optimizer with default parameters ($\beta_1 = 0.9, \beta_2 = 0.999$) and with weight decay modification \cite{loshchilov2017decoupled} and the cross-entropy loss function. We replace numbers, emails and links with generic tokens, we tokenize input text with the TokTok tokenizer\footnote{\url{https://github.com/jonsafari/tok-tok}} and we use a customized stemmer\footnote{\url{https://github.com/UFAL-DSG/alex/blob/master/alex/utils/czech_stemmer.py}}.

We use different hyper-parameters for each dataset, see Appendix \ref{sec:apendix-hyper} for the complete settings.

% https://towardsdatascience.com/multilingual-transformers-ae917b36034d Tady je jejich shrnuti

\subsection{Transformer Models}
\textcolor{black}{In total, we use eight different transformer-based models (five of them are multilingual). All of them are based on the original BERT model. They use only the encoder part of the original Transformer
%\footnote{Please, see the cited papers for details.}
\cite{attention-all-transformer}, although their pre-training procedure may differ.} There are also \textit{text-to-text} models like T5 \cite{2020-t5} and BART \cite{lewis2019bart} and their multilingual versions mT5 \cite{xue2020-mt5} and mBART \cite{liu2020multilingual,tang2020multilingual-mbart50}. 
%, respectively. 
The main difference from BERT-like models is that they use the full encoder-decoder architecture of the Transformer. They are mainly intended for text generation tasks (e.g., abstractive summarization). %We did not use them and 
We decided to use only the BERT-like models with the same architecture because they fit more for the classification task.

% \todo{Pokud by zbyl cas tak sem popsat ty tasky}
\par All the models are pre-trained on a modified language modeling task, for example, \textit{Masked  Language  Modeling} (MLM) and eventually on some classification task like \textit{Next Sentence Prediction} (NSP) or \textit{Sentence Ordering Prediction} (SOP), see \cite{devlin-etal-2019-bert, lan2019albert} for details. The evaluated models differ in the number of parameters (see Table \ref{tab:models-size}) and thus, their performance is also 
very different, see Section \ref{sec:results}.

\begin{table}[ht!]
\catcode`\-=12
\begin{adjustbox}{width=0.9\linewidth,center}
\begin{tabular}{lrrr} 
\toprule
Model        &  \multicolumn{1}{c}{\#Params} & \multicolumn{1}{c}{Vocab} & \multicolumn{1}{c}{\#Langs} \\ \midrule
Czert-B      & 110M       & 30k   & 1                             \\
Czert-A      & 12M        & 30k   & 1                             \\
RandomALBERT & 12M        & 30k   & 1                             \\
mBERT        & 177M       & 120k  & 104                           \\
SlavicBERT   & 177M       & 120k  & 4                             \\
XLM          & 570M       & 200k  & 100                           \\
XLM-R-Base   & 270M       & 250k  & 100                           \\
XLM-R-Large  & 559M       & 250k  & 100                          \\
\bottomrule
\end{tabular}
\end{adjustbox}
\caption{Models statistics with a number of parameters, vocabulary size and a number of supported languages.}
\label{tab:models-size}
\end{table}

\paragraph{Czert-B} is Czech version of the of the original BERT\textsubscript{BASE} model \cite{devlin-etal-2019-bert}. The only difference is that during the pre-training, the authors increased the batch size to 2048 and they slightly modified the NSP prediction task \cite{sido2021czert}.

\paragraph{Czert-A} is the Czech version of the ALBERT model \cite{lan2019albert}, also with the same modification as \texttt{Czert-B}, i.e., batch size was set to 2048 and the modified NSP prediction task is used instead of the SOP task \cite{sido2021czert}. 

\paragraph{RandomALBERT} we follow the evaluation in \cite{sido2021czert} and we also test randomly initialized ALBERT model without any pre-training to show the importance of pre-training of such models and its performance influence on the polarity detection task.

\paragraph{mBERT} \cite{devlin-etal-2019-bert} is a multilingual version of the original BERT\textsubscript{BASE}, jointly trained on 104 languages. 

\paragraph{SlavicBERT} \cite{arkhipov2019tuning-SlavicBert} is initialized from the \texttt{mBERT} checkpoint and further pre-trained with a modified vocabulary only for four Slavic languages (Bulgarian, Czech, Polish and Russian).

\paragraph{XLM} \cite{lample2019cross-xlm} utilizes the training procedure of the original BERT model for multilingual settings mainly by using the Byte-Pair Encoding (BPE) and increasing the shared vocabulary between languages. 

% ======== Poznamky
% precist XLM clanek Crosslingual language model pretraining
% https://arxiv.org/pdf/1901.07291.pdf
% https://papers.nips.cc/paper/2019/hash/c04c19c2c2474dbf5f7ac4372c5b9af1-Abstract.html
% https://towardsdatascience.com/xlm-enhancing-bert-for-cross-lingual-language-model-5aeed9e6f14b#:~:text=A%20new%20paper%20by%20Facebook,between%20words%20in%20different%20languages.
% ==================
% TODO v related work maji odkaz na zero-shot cross-lingual classification

\paragraph{XLM-R-Base} \cite{xlm-r} is a multilingual version of the RoBERTa \cite{liu2019roberta} specifically optimized and pre-trained for 100 languages.

% ======== Poznamky
% on 2.5 TB data
% roberta clanek Unsupervised Cross-lingual Representation Learning at Scale
% https://arxiv.org/abs/1911.02116
% https://www.aclweb.org/anthology/2020.acl-main.747.pdf
% https://arxiv.org/pdf/2105.00572.pdf precist XLM-R XXL
% https://arxiv.org/abs/1907.11692 roberta clanek - precist nejdrive
% ==================

\paragraph{XLM-R-Large} \cite{xlm-r} is the same model as the \texttt{XLM-R-Base}, but it is larger (it has more parameters).

% ======== Poznamky
% XLM-R clanek
% https://arxiv.org/pdf/1911.02116.pdf
% Roberta clanek
% hlavni zmeny oproti bertu
% 1) trenujou model dele, s vetsim batchem a pres vic dat
% 2) odstranili next sentence prediction
% 3) trenujou na delsi sekvenci
% 4) dynamicky meni maskovaci pattern
% ==================

\subsection{Transformers Fine-Tuning}
To utilize the models for text classification, we follow the default approaches mentioned in the corresponding models' papers and we fine-tune all parameters of the models. In all models except \texttt{XLM}, we use the final hidden vector $\vec{h} \in \mathbb{R}^H$ of the special classification token \texttt{[CLS]} or \texttt{<s>} taken from the pooling layer\footnote{The pooling layer is a fully-connected layer of size $H$ with a hyperbolic tangent used as the activation function.} of BERT or RoBERTa models, respectively. The vector $\vec{h}$ represents the entire encoded sequence input, where $H$ is the hidden size of the corresponding model. We add a task-specific linear layer (with a dropout set to $0.1$) represented by a matrix $\vec{W} \in \mathbb{R}^{|C| \times H}$, where C is a set of classes. We compute the classification output, i.e., the input sample being classified as  class  $c \in C$ as $c = \argmax(\vec{h}\vec{W}^{T})$.

% We compute the classification output, i.e., the probability $p(c \mid \vec{h})$ of a sample being classified with a label $c \in C$ with a softmax function as follows: $p(c \mid \vec{h}) = \texttt{softmax}(\vec{h}\vec{W}^{T})$. 
% ja tam nemam na konci softmax ten je jenom v tej Loss - ja delam argmax

\par In the case of the \texttt{XLM} model, we take the last hidden state (without any pooling layer) of the first input token and we apply the same linear layer ($\vec{W} \in \mathbb{R}^{|C| \times H}$) and approach to obtain the classification output. For learning, we use the Adam optimizer with default parameters and with weight decay (same as for the \texttt{LSTM} model), and the cross-entropy loss function. See Section \ref{sec:monolingual-experiments} and Appendix \ref{sec:transformer-appendix} for the hyper-parameters we used.

% ======== Poznamky
% https://huggingface.co/transformers/_modules/transformers/modeling_utils.html - sequence summary
% melo by to byt stejne vezme to prvni hidden state a posle to pres linear vrstvu s dropoutem 0.1
% todo podivat se jak to ma XLM a XLM-R
% U xlm to vypada ze tam pridaji linear vrstvu a berou 
% https://huggingface.co/transformers/_modules/transformers/models/xlm/modeling_xlm.html#XLMForSequenceClassification

% Nemazat!!!
% Model(vystup) se dela tak ze se vezme vystup pro prvni token (tj. ten klasifikacni a ten se prozene Linearni vrstvou a za tou je hyperbolickej tangens (Trdia BertPooler), BertModel pak vrati, tenhle vystup + ten originalni co sel do linear vrstvy a tanh.
% BertForSequenceClassification pak vezme ten vystup prohnanej tanh a prozene da na nej dropout a pak ho prozene linearni vrstvou bez aktivace 
% ==================

\section{Experiments \& Results}
\label{sec:results}
We perform two types of experiments, i.e., \textit{monolingual} and \textit{cross-lingual}. In \textit{monolingual} experiments, we fine-tune and evaluate the Transformer models for each dataset separately on three-class (\textit{positive, negative} and \textit{neutral}) and two-class (\textit{positive} and \textit{negative}) sentiment analysis. We also implemented the logistic regression (\texttt{lrc}) and \texttt{LSTM} baseline models and we compare the results with the existing works.

\par In \textit{cross-lingual} experiments, we test the ability of four multilingual transformer-based models to transfer knowledge between English and Czech. We run the multilingual models only on the two-class datasets (\textit{positive} and \textit{negative}). We fine-tune either on English (IMDB) or Czech (CSFD), and then we evaluate on the other language. Thus we perform the \textit{zero-shot cross-lingual} classification. We decided to use the IMDB and CSFD dataset because they are from the same domain i.e., movie reviews.

\par Each experiment\footnote{Except for the experiments with the \texttt{lrc} model.} was performed at least five times and we report the results using the macro $F_1$ score.

\begin{table*}[ht!]
\catcode`\-=12
\begin{adjustbox}{width=\linewidth,center}
\begin{tabular}{llllcccc}
\toprule
 \multirow{2}{*}{Model}               & \multicolumn{3}{c}{3 Classes}                                                                                            &       & \multicolumn{3}{c}{2 Classes}                                                                                                      \\ \cline{2-4} \cline{6-8}
                & \multicolumn{1}{c}{CSFD}                 & \multicolumn{1}{c}{FB}                   & \multicolumn{1}{c}{Mallcz}    &            & \multicolumn{1}{c}{CSFD}                   & \multicolumn{1}{c}{FB}                    & \multicolumn{1}{c}{Mallcz}                \\ \midrule
lrc  (ours)     & \multicolumn{1}{c}{79.63}               & \multicolumn{1}{c}{67.86}               & \multicolumn{1}{c}{76.71}         &       & \multicolumn{1}{c}{91.42}                 & \multicolumn{1}{c}{88.12}                & \multicolumn{1}{c}{88.98}                \\
LSTM   (ours)          &     79.88 $\pm$ 0.18                                     & 72.89 $\pm$ 0.49                                        &     73.43 $\pm$ 0.12                           &           &   91.82 $\pm$ 0.09                                          &    90.13 $\pm$ 0.17                                     &    88.02 $\pm$ 0.24                                       \\ \hdashline
Czert-A     & $79.89 \pm 0.60$            & $73.06 \pm 0.59$             & $76.79 \pm 0.38$     &        & $91.84 \pm 0.84$            & $91.28 \pm 0.18$        & 91.20 $\pm$ 0.26        \\
Czert-B        & 84.80 $\pm$ 0.10         & 76.90 $\pm$ 0.38         & 79.35 $\pm$ 0.24        &    & 94.42 $\pm$ 0.15           & 93.97 $\pm$ 0.30           & 92.87 $\pm$ 0.15            \\
mBERT           & 82.74 $\pm$ 0.16         & 71.61 $\pm$ 0.13        & 70.79 $\pm$ 5.74   &         & 93.11 $\pm$ 0.29       & 88.76 $\pm$ 0.42           & 72.79 $\pm$ 3.09            \\
SlavicBERT      & 82.59 $\pm$ 0.12         & 73.93 $\pm$ 0.53          & 75.34 $\pm$ 2.54     &      & 93.47 $\pm$ 0.33          & 89.84 $\pm$ 0.43          & 90.99 $\pm$ 0.15      \\
RandomALBERT   & 75.79 $\pm$ 0.18           & 62.53 $\pm$ 0.46          & 64.81 $\pm$ 0.25      &     & 89.99 $\pm$ 0.21           & 81.71 $\pm$ 0.56          & 85.38 $\pm$ 0.10       \\
XLM-R-Base      & 84.82 $\pm$ 0.10         & 77.81 $\pm$ 0.50        & 75.43 $\pm$ 0.07      &   & 94.32 $\pm$ 0.34          & 93.26 $\pm$ 0.74         & 92.56 $\pm$ 0.07         \\
XLM-R-Large     & \textbf{87.08 $\pm$ 0.11}& \textbf{81.70 $\pm$ 0.64}  & \textbf{79.81 $\pm$ 0.21}  & & \textbf{96.00 $\pm$ 0.02} & \textbf{96.05 $\pm$ 0.01}   & \textbf{94.37 $\pm$ 0.02}  \\
XLM             & 83.67 $\pm$ 0.11           & 71.46 $\pm$ 1.58          & 77.56 $\pm$ 0.08       &   & 93.86 $\pm$ 0.18         & 89.94 $\pm$ 0.27           & 91.97 $\pm$ 0.22          \\ \hdashline
\cite{habernal-etal-2013-sentiment}$\dagger$  & \multicolumn{1}{c}{79.00}                & \multicolumn{1}{c}{69.00}                & \multicolumn{1}{c}{75.00}   &              & \multicolumn{1}{c}{-}                      & \multicolumn{1}{c}{90.00}                 & \multicolumn{1}{c}{-}                     \\
\cite{brychcin-habernal-2013-unsupervised-sentiment}$\dagger$     & \multicolumn{1}{c}{81.53 $\pm$ 0.30}       & \multicolumn{1}{c}{-}                    & \multicolumn{1}{c}{-}      &               & \multicolumn{1}{c}{-}                      & \multicolumn{1}{c}{-}                     &      \multicolumn{1}{c}{-}                                      \\
% \cite{SLON-Lenc2016Neural}     & \multicolumn{1}{c}{71.00}                & \multicolumn{1}{c}{69.40}                & \multicolumn{1}{c}{75.50}       &          & \multicolumn{1}{c}{-}                      & \multicolumn{1}{c}{-}                     & \multicolumn{1}{c}{-}                     \\
% \cite{sido-curic}      & \multicolumn{1}{c}{80.50 $\pm$ 0.16}         & \multicolumn{1}{c}{-}                    & \multicolumn{1}{c}{-}       &              & \multicolumn{1}{c}{-}                      & \multicolumn{1}{c}{-}                     &                                           \\
\cite{libovicky2018solving}*  & \multicolumn{1}{c}{80.80 $\pm$ 0.10}        & \multicolumn{1}{c}{-}                    & \multicolumn{1}{c}{-}       &              & \multicolumn{1}{c}{-}                      & \multicolumn{1}{c}{-}                     & \multicolumn{1}{c}{-}                     \\
\cite{kybernetika-bert}*  & \multicolumn{1}{c}{-}                    & \multicolumn{1}{c}{-}                    & \multicolumn{1}{c}{-}            &         & \multicolumn{1}{c}{93.80}                   & \multicolumn{1}{c}{-}                     & \multicolumn{1}{c}{-}       \\ \bottomrule             
\end{tabular}
\end{adjustbox}
\caption{The final monolingual results as macro $ F_{1}$ score for all three Czech polarity datasets on two and three classes. For experiments with neural networks performed by us, we present the results with a 95\% confidence interval. The models from papers marked with $\dagger$ were evaluated with 10-fold cross-validation and the ones marked with * were evaluated on custom data split.} \label{tab:monolingual-res}
\end{table*}

\subsection{Monolingual Experiments}
\label{sec:monolingual-experiments}
The goal of the monolingual experiments is to reveal the current state-of-the-art performance on the Czech polarity datasets, namely CSFD, FB and Mallcz (see Section \ref{sec:data}) and provide a comparison between the available models and their settings.

\par As we already mentioned, we split the datasets into training, development and testing parts. There is no official split for the datasets and we found out that all the available works usually use either 10-fold cross-validation or they split\footnote{The authors do not provide any recipe to reproduce the results.} the datasets on their own, the $\dagger$ and * symbols in Table \ref{tab:monolingual-res}, respectively causing the comparison to be difficult.

\par We fine-tune all models on training data and we measure the results on the development data. We select the model with the best performance on the development data and we fine-tune the model on combined training and development data. We report the results in Table \ref{tab:monolingual-res} on the testing data with 95\% confidence intervals.

\par Firstly, we re-implemented the logistic regression classifier (\texttt{lrc}) with the best feature combination from \cite{habernal-etal-2013-sentiment} and we report the results 
%in Table \ref{tab:monolingual-res} 
on our data split. We can see that we obtained very similar results to the ones stated in \cite{habernal-etal-2013-sentiment}. We also tried to improve this baseline with Tf-idf weighting, but it did not lead to any significant improvements, so we decided to keep the settings the same as in \cite{habernal-etal-2013-sentiment}, so the results are comparable.

\par For the \texttt{LSTM} model, we tried different combinations of hyper-parameters (learning rate, optimizer, dropout, etc.). We report the used hyper-parameters for the results from Table \ref{tab:monolingual-res} in Appendix \ref{sec:transformer-appendix}. 
%As is clear from the results, 
Our implementation is only about 1 \% worse than LSTM with the self-attention model from \cite{libovicky2018solving}, but they used a different data split. For the Mallcz dataset, we were not able to outperform the \texttt{lrc} baseline with the \texttt{LSTM} model. 

\par
\begin{figure}%
    \centering
    \subfloat[\centering CSFD -- Czert-B]{{\includegraphics[width=7.cm]{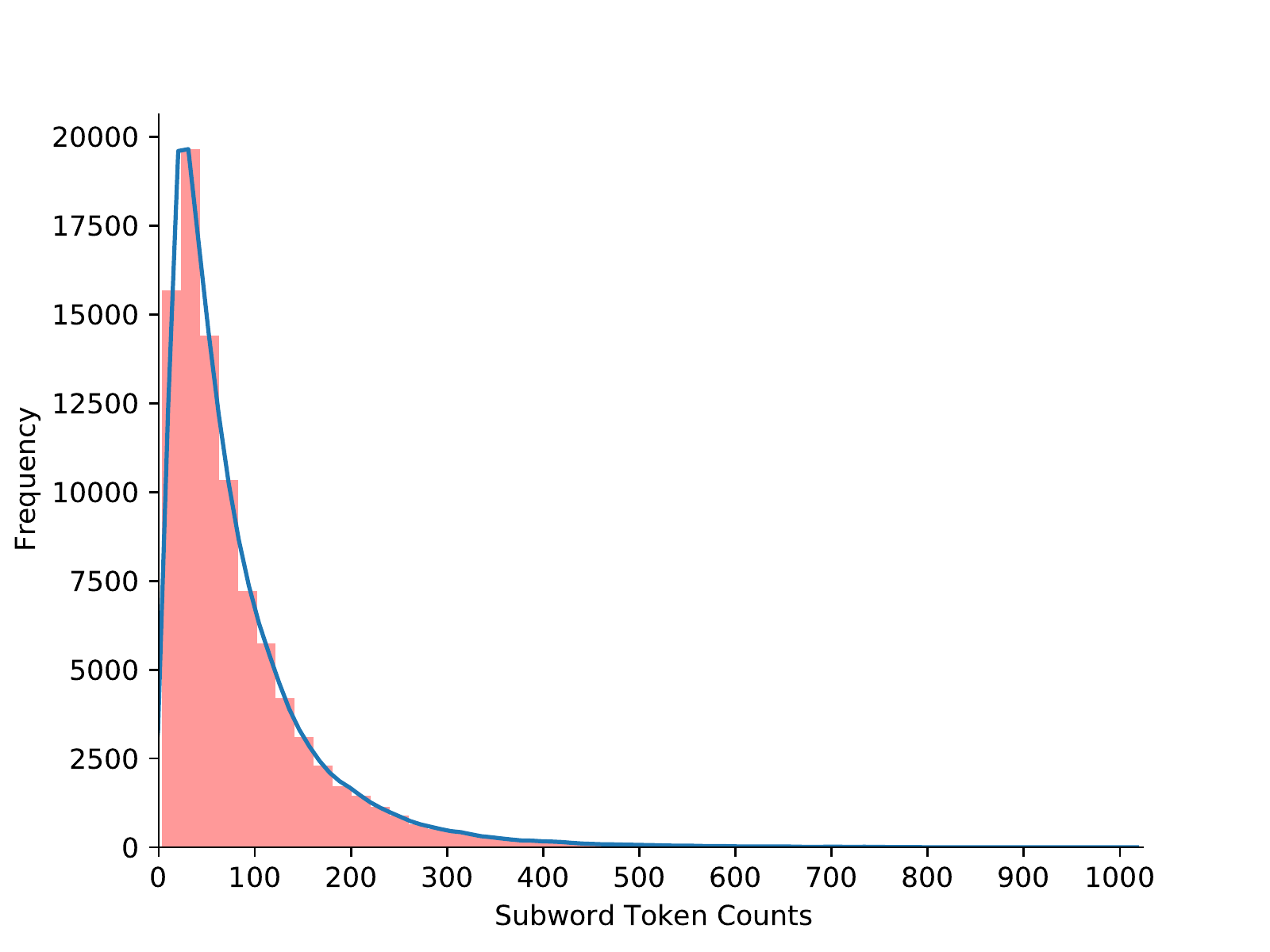} }}%
    \qquad
    \subfloat[\centering Mallcz -- Czert-B]{{\includegraphics[width=7cm]{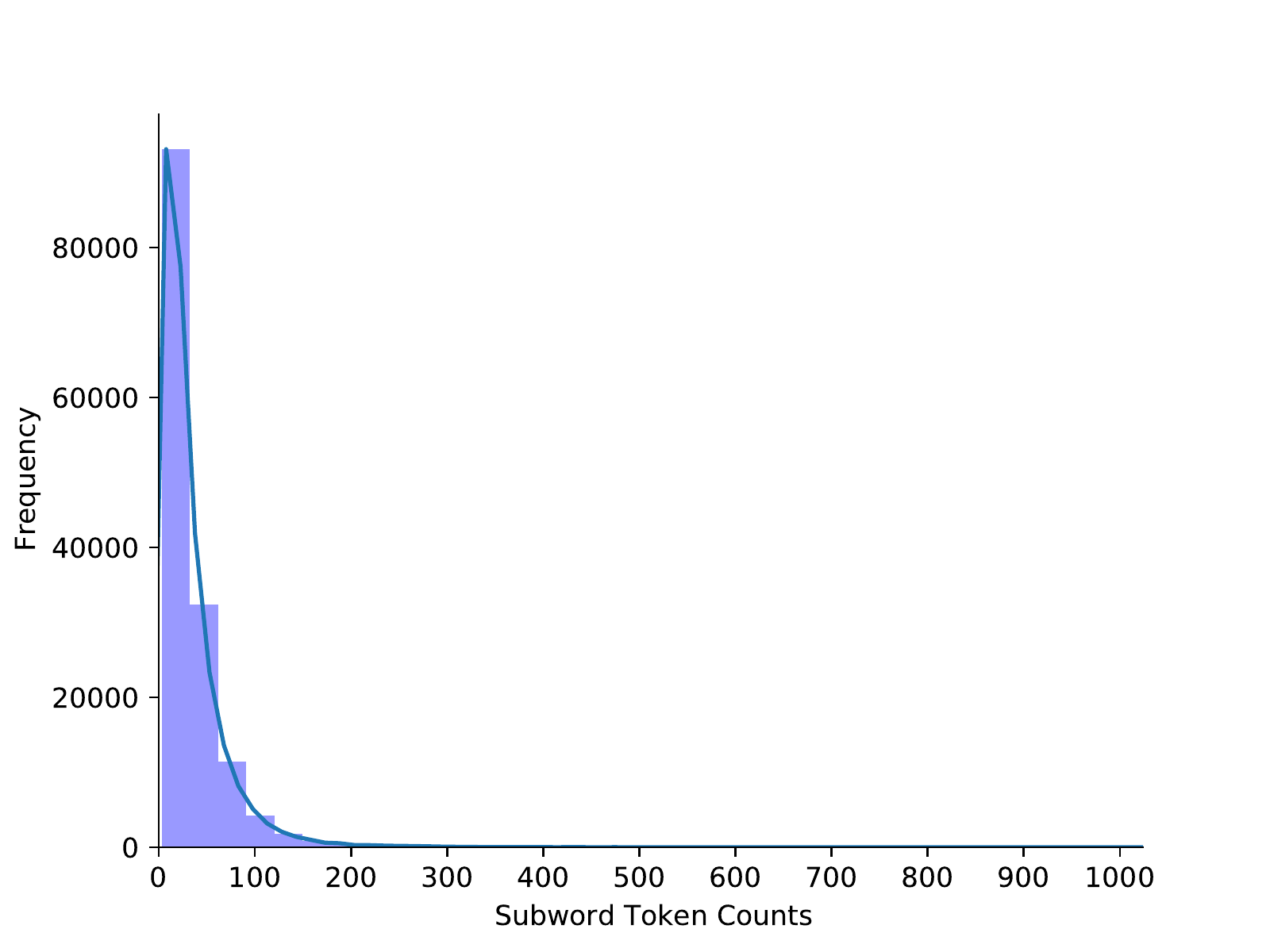} }}%
    \caption{Subword token histograms for the CSFD and Mallcz datasets for the \texttt{Czert-B} model.}%
    \label{fig:czert-B-subwords}%
\end{figure}

\begin{figure}%
    \centering
    \subfloat[\centering CSFD -- XLM-R-Base and XLM-R-Large]{{\includegraphics[width=7.cm]{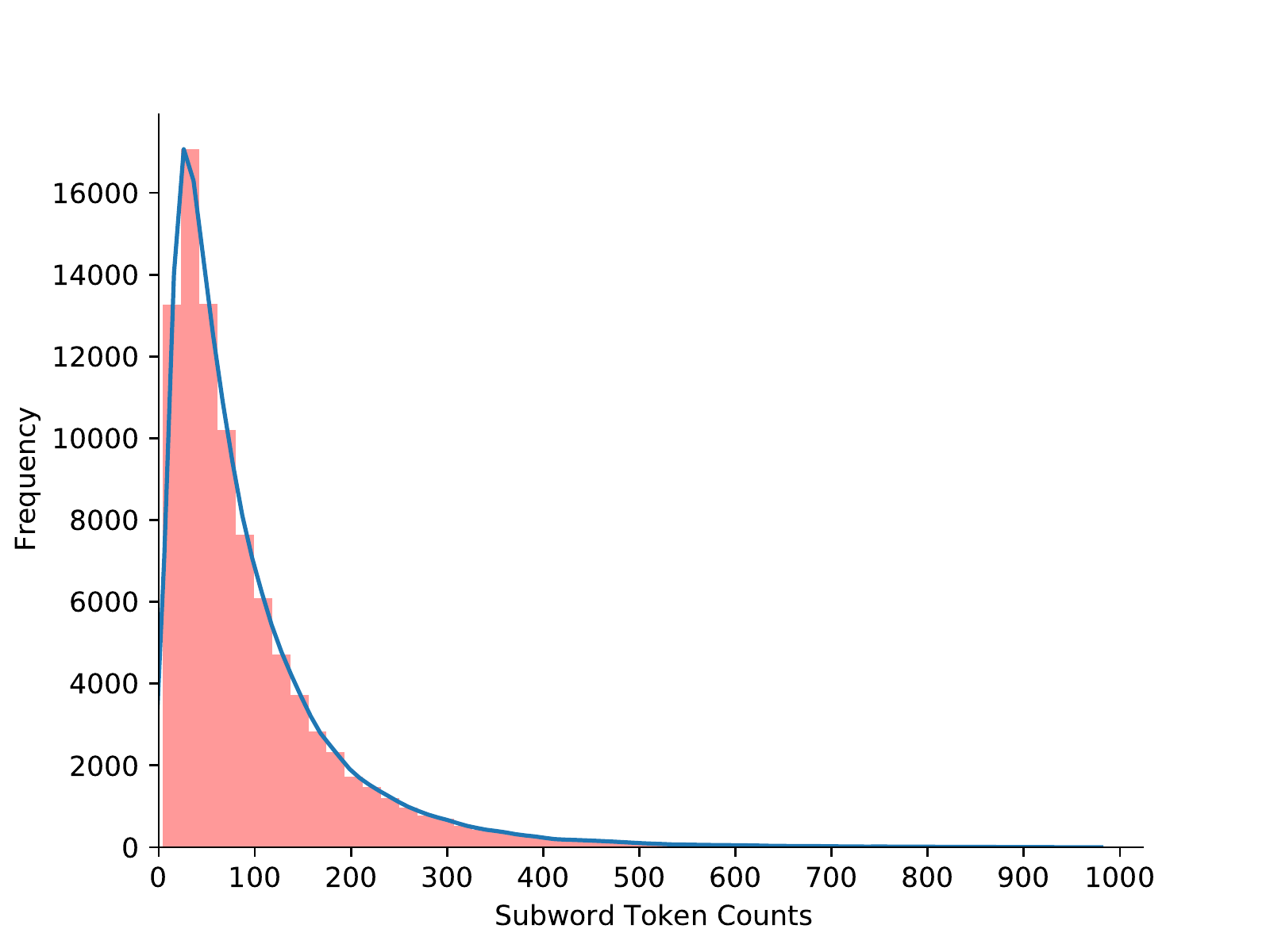} }}%
    \qquad
    \subfloat[\centering Mallcz -- XLM-R-Base and XLM-R-Large]{{\includegraphics[width=7cm]{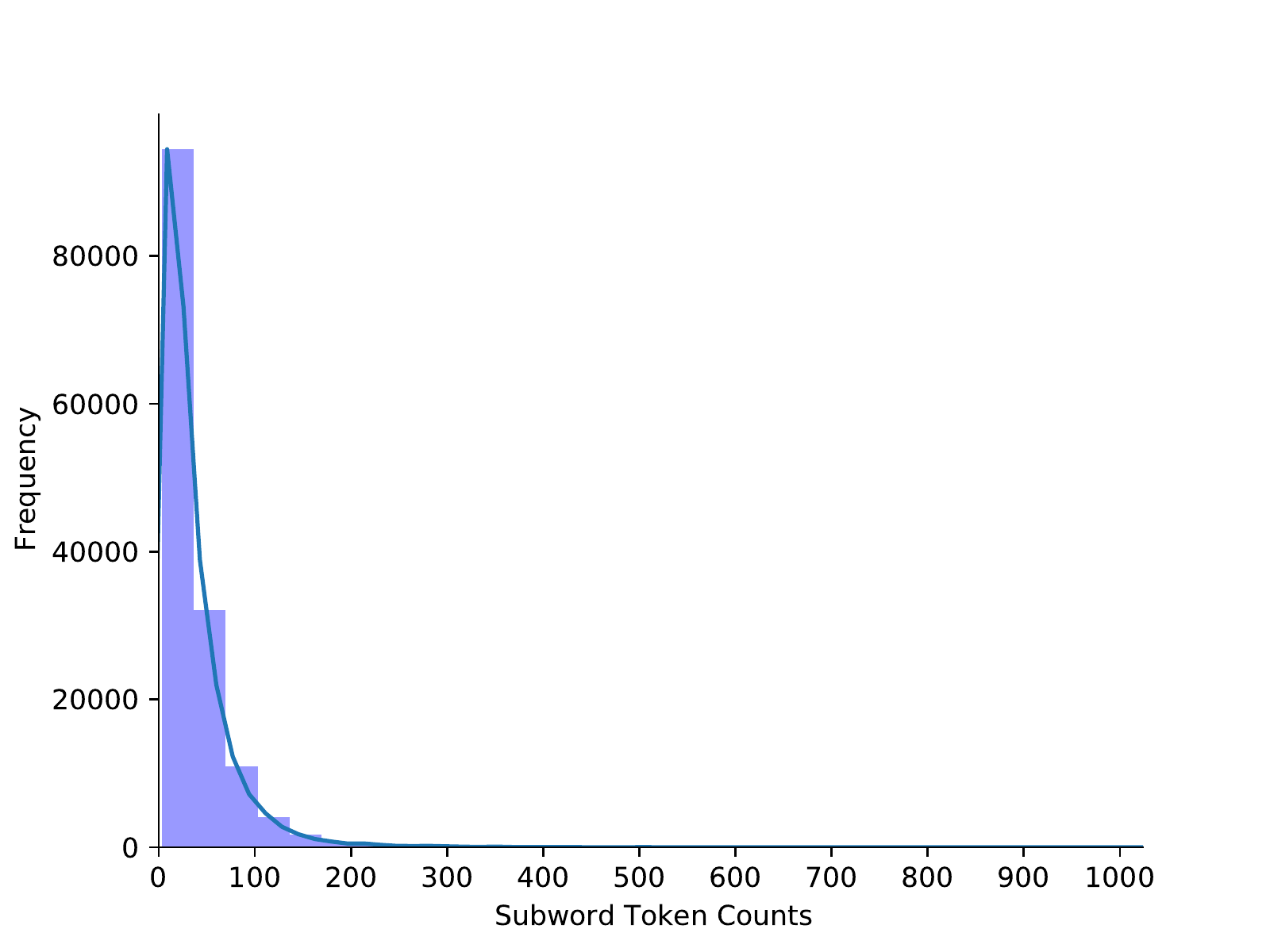} }}%
    \caption{Subword token histograms for the CSFD and Mallcz datasets for the \texttt{XLM-R-Base} and \texttt{XLM-R-Large} models.}%
    \label{fig:xlm-r-subwords}%
\end{figure}

\begin{figure}%
    \centering
    \subfloat[\centering CSFD -- mBERT]{{\includegraphics[width=7.cm]{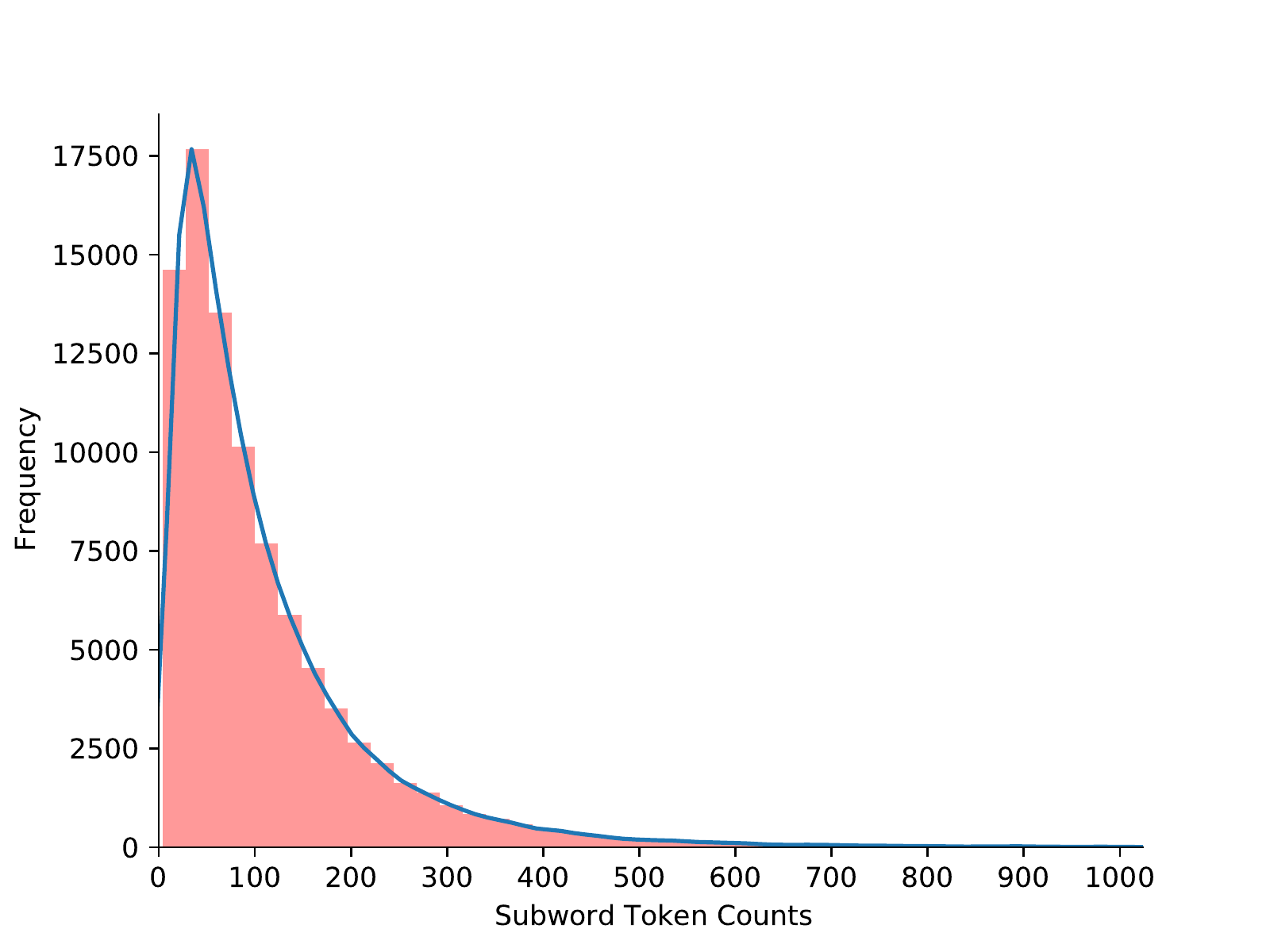} }}%
    \qquad
    \subfloat[\centering Mallcz -- mBERT]{{\includegraphics[width=7cm]{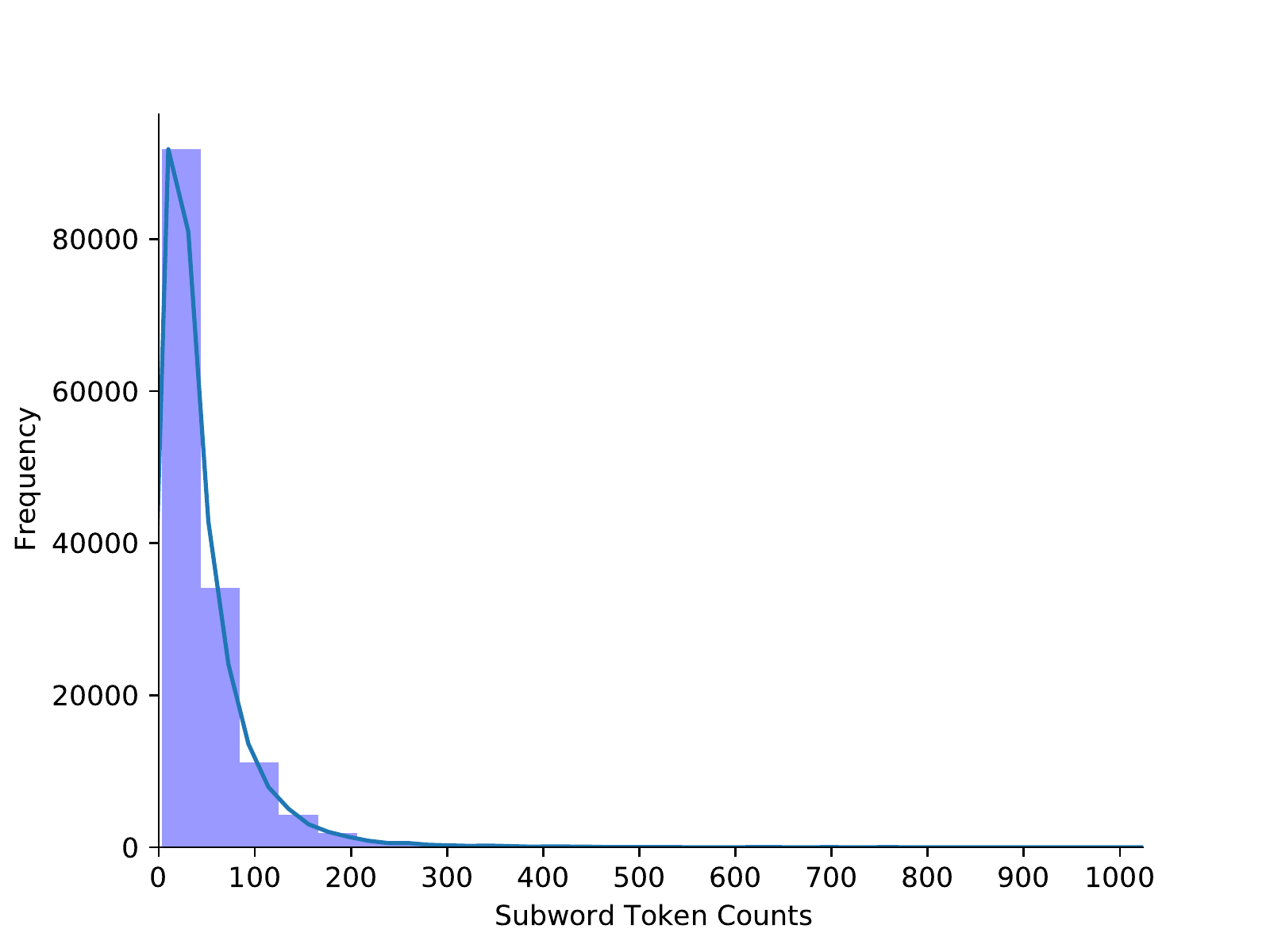} }}%
    \caption{Subword token histograms for the CSFD and Mallcz datasets for the \texttt{mBERT} model.}%
    \label{fig:mbert-subwords}%
\end{figure}

 We fine-tune all parameters of the seven pre-trained BERT-based models and one randomly initialized ALBERT model. In our experiments, we use constant learning rate and also linear learning rate decay (without learning rate warm-up) with the following initial learning rates: 2e-6, 2e-5 and 2.5e-5. We got inspired by the ones used in \cite{sun2019fine}. Based on the average number of tokens for each dataset and models' tokenizer (see Table \ref{tab:models-tokenizer-size} \textcolor{black}{and Figures \ref{fig:czert-B-subwords}, \ref{fig:xlm-r-subwords}, \ref{fig:mbert-subwords})}\footnote{\textcolor{black}{The distributions of the other models were similar to those shown in the mentioned Figures.}}, we use a max sequence length of 64 and a batch size of 32 for the FB dataset. We restrict the max sequence length for the CSFD and Mallcz datasets to 512 and use a batch size of 32. All other hyper-parameters of the models are set to the pre-trained models' defaults. See Table \ref{tab:transformer-hyperparameters} in Appendix \ref{sec:transformer-appendix} for the reported results' hyper-parameters.

\begin{table}[ht!]
\catcode`\-=12
\begin{adjustbox}{width=\linewidth,center}
\begin{tabular}{lcccccccc} \toprule
    \multirow{2}{*}{Model}             & \multicolumn{2}{c}{CSFD}           &             & \multicolumn{2}{c}{FB}    &                            & \multicolumn{2}{c}{Mallcz}                      \\ \cline{2-3} \cline{5-6} \cline{8-9}
              & Avg.               & \multicolumn{1}{c}{Max.} & & \multicolumn{1}{c}{Avg.} & \multicolumn{1}{c}{Max.} && Avg.               & \multicolumn{1}{c}{Max.}  \\ \midrule

Czert-B     & 84.5                  & 1000 &                   & 20.3                        & 64  &                    & 34.3                  & 1471                    \\
mBERT         & 111.6                 & 1206      &              & 25.6                        & 66     &                 & 46.6                  & 2038                    \\
SlavicBERT    & 83.6                  & 983     &                & 20.7                        & 62    &                  & 34.3                  & 1412                    \\

XLM           & 100.5                 & 1058  &                  & 22.6                        & 64    &                  & 41.0                  & 1812                   \\ \hdashline

Czert-A   & \multirow{2}{*}{81.7} & \multirow{2}{*}{993}  &  & \multirow{2}{*}{19.7}       & \multirow{2}{*}{62}  &   & \multirow{2}{*}{32.6} & \multirow{2}{*}{1435}   \\
RandomALBERT &                       &                         &                             &                         &                       &                         \\
\hdashline
XLM-R-Base    & \multirow{2}{*}{93.9} & \multirow{2}{*}{952} &    & \multirow{2}{*}{20.4}       & \multirow{2}{*}{53} &   & \multirow{2}{*}{37.5} & \multirow{2}{*}{1670}   \\
XLM-R-Large   &                       &                         &                             &                         &                       &                         \\ \bottomrule
\end{tabular}
\end{adjustbox} \caption{The average and maximum number of sub-word tokens for each model's tokenizer and dataset.} \label{tab:models-tokenizer-size}
\end{table}

\par We repeated all the basic experiments with the polarity detection task from \cite{sido2021czert} with the new data split. Our results do not significantly differ, as shown in Table \ref{tab:czert-res-comparison} and in Appendix \ref{sec:transformer-appendix}. If we compare the BERT model from \cite{kybernetika-bert} with the \texttt{Czert-B, mBERT} and \texttt{SlavicBERT} models\footnote{All of them should have the same or almost the same architecture and a similar number of parameters.}, we can see that on the binary task, they also perform very similarly, i.e., around 93 \%, but again they used different test data (the entire CSFD dataset\footnote{The examples with positive and negative classes.}). The obvious observation is that the \texttt{XLM-R-Large} model is superior to all others by a significant margin for any dataset. Only for the three-class Mallcz dataset, the \texttt{Czert-B} model is competitive (the confidence intervals almost overlap).
From the results for the \texttt{RandomALBERT} model, we can see how important is the pre-training phase for Transformers, since the model is even worse than the logistic regression classifier\footnote{The model was trained for a maximum of 15 epochs and it would probably get better with a higher number of epochs, but the other models were trained for the same or lower number of epochs.}.

\subsection{Cross-lingual Experiments}
The cross-lingual experiments were performed with the multilingual models that support English and Czech. For these experiments, we use linear learning rate decay with an initial learning rate of 2e-6.

\par Firstly, we fine-tuned the models on the English IMDB dataset and we evaluated them on the test part of the Czech binary CSFD dataset (i.e., zero-shot cross-lingual classification). We randomly selected 5k examples from the IMDB dataset as the development data. The rest of the 45k examples is used as training data. We select the models that perform best on the English development data\footnote{The \textit{dev (en)} column in Table \ref{tab:cross-en-cs}.} and we report the results in Table \ref{tab:cross-en-cs}. The \textit{test (cs)} column refers to results obtained on the CSFD testing part. For easier comparison, we also include the \textit{Monoling. (cs)} column that contains the results (same as in Table \ref{tab:monolingual-res}) for models trained on Czech data. The \texttt{XLM-R-Large} was able to achieve results only about $4.4$ \% worse than the same model that was fine-tuned on Czech data. It is a great result if we consider that the model has never seen any labeled Czech data. The \texttt{XLM} and \texttt{mBERT} models perform much worse.

\begin{table}[ht!]
\catcode`\-=12
\begin{adjustbox}{width=\linewidth,center}
\begin{tabular}{lcccc} \toprule
     \multirow{2}{*}{Model}        & \multicolumn{2}{c}{EN $\rightarrow$ CS} & & Monoling. (cs)   \\  \cline{2-3} \cline{5-5}
            & dev (en)           & test (cs)          &  &   \\ \midrule
XLM-R-Base  & 94.52 $\pm$ 0.12   & 88.01 $\pm$ 0.28    &  & 94.32 $\pm$ 0.34 \\
XLM-R-Large & 95.86 $\pm$ 0.06   & \textbf{91.61 $\pm$ 0.06}   &  & 96.00 $\pm$ 0.02 \\
XLM         & 92.76 $\pm$ 0.34   & 75.37 $\pm$ 0.29   &  & 93.86 $\pm$ 0.18 \\
mBERT       & 93.07 $\pm$ 0.03    & 76.32 $\pm$ 1.13    &  & 93.11 $\pm$ 0.29 \\ \bottomrule
\end{tabular}
\end{adjustbox}
\caption{Macro $F_1$ score for cross-lingual experiments from English to Czech.} \label{tab:cross-en-cs}
\end{table}

The second type of cross-lingual experiment was performed in a reverse direction, i.e., from Czech to English. We use the Czech CSFD training and testing data for fine-tuning and we evaluate the model on the English IMDB test data. We report the results in Table \ref{tab:cross-cs-en} using the accuracy because the current state-of-the-art works \cite{thongtan-phienthrakul-2019-sentiment,sun2019fine} use this metric. Similarly to the previous case, we selected the model that performs best on Czech CSFD development data. For these experiments, the \texttt{mBERT} did not converge. As in the previous experiment, the \texttt{XLM-R-Large} performs best and it achieves almost $94$ \% accuracy that is only $3.4$ \% below the current SotA result from \cite{thongtan-phienthrakul-2019-sentiment}.

\begin{table}[ht!]
\catcode`\-=12
\begin{adjustbox}{width=\linewidth,center}
\begin{tabular}{lcc} \toprule
  \multirow{2}{*}{Model}           & \multicolumn{2}{c}{CS $\rightarrow$ EN} \\
            \cline{2-3}
            & dev (cs)                & test (en)              \\ \midrule
XLM-R-Base  & 94.22 $\pm$ 0.01        & 89.53 $\pm$ 0.15       \\
XLM-R-Large & 95.65 $\pm$ 0.17        & \textbf{93.98 $\pm$ 0.10}       \\
XLM         & 93.66 $\pm$ 0.13         & 78.24 $\pm$ 0.46         \\ \hdashline
% mBERT       & -                       & -                      \\ \hdashline
\cite{thongtan-phienthrakul-2019-sentiment}    & -                       &  97.42                     \\
\cite{sun2019fine}   & -                       &     95.79                  \\ \bottomrule
\end{tabular}
\end{adjustbox}
\caption{Accuracy results for cross-lingual experiments from Czech to English.} \label{tab:cross-cs-en}
\end{table}

\par Based on the results, we can conclude that the \texttt{XLM-R-Large} model is very capable of transferring knowledge between English and Czech (and probably between other languages as well). It is also important to note that Czech and English are languages from a different language family with a high number of differences both in syntax and grammar.

\subsection{Discussion \& Remarks}
We can see from the results that the recent pre-trained transformer-based models beat the older approaches (\texttt{lrc} and \texttt{LSTM}) by a large margin. The monolingual \texttt{Czert-B} model is in general outperformed only by the \texttt{XLM-R-Large} and \texttt{XLM-R-Base} models, but these models have five times/three times more parameters, and eight times larger vocabulary. Taking into account these facts, the \texttt{Czert-B} model is still very competitive. It may be beneficial in certain situations to use a smaller model like this that does not need such computational resources as the ones that are required by the \texttt{XLM-R-Large}.

\par During the fine-tuning, we observed that in most cases, the lower learning rate 2e-6 (see Table \ref{tab:transformer-hyperparameters} in Appendix \ref{sec:transformer-appendix}) leads to better results. Thus we recommend using the same one or similar order. The higher learning rates tend to provide worse results and the model does not converge.

\par According to the generally higher confidence interval, the fine-tuning of a smaller dataset like FB that has only about 6k training examples is, in general, less stable and more prone to overfitting than training a model on datasets with tens of thousands of examples. We also noticed that fine-tuning of the \texttt{mBERT} and \texttt{SlavicBERT} on the Mallcz dataset is very unstable (see the confidence interval in Table \ref{tab:monolingual-res}). Unfortunately, we did not find out the reason. A more detailed error analysis could reveal the reason.

\section{Conclusion}
In this work, we %fine-tuned and 
evaluated the performance of available transformer-based models for the Czech language on the task of polarity detection. We compared the performance of the monolingual and multilingual models and we showed that the large \texttt{XLM-R-Large} model can outperform the monolingual \texttt{Czert-B} model. The older approach based on recurrent neural networks is surpassed by a very large margin by the Transformers. Moreover, we achieved new state-of-the-art results on all three Czech polarity detection datasets.

\par We performed zero-shot cross-lingual polarity detection from English to Czech (and vice versa) with four multilingual models. We showed that the \texttt{XLM-R-Large} is able to detect polarity in another language without any labeled data. The model performs no worse than 4.4 \% in comparison to our new state-of-the-art monolingual model. To the best of our knowledge, this is the first work that aims at cross-lingual polarity detection in Czech. Our code and pre-trained models are publicly available.

% zapojit vic dat z ostatnich jazyku jestli to bude stale jeste zlepsovat a jestli to bude fungovat i v dalsich domenach

\par \textcolor{black}{In the future work, we intend to perform a deep error analysis to find in which cases the current models fail and compare approaches that use the linear cross-lingual transformations \cite{artetxe-etal-2018-robust,brychcin2020linear} that explicitly map semantic spaces into one shared space. The second step in the cross-lingual settings is to employ more than two languages and utilize the models for different domains.}

\section*{Acknowledgments}
This work has been partly supported by ERDF ”Research and Development of Intelligent Components of Advanced Technologies for the Pilsen Metropolitan Area (InteCom)” (no.: CZ.02.1.01/0.0/0.0/17 048/0007267); and by Grant No. SGS-2019-018 Processing of heterogeneous data and its specialized applications. Computational resources were supplied by the project "e-Infrastruktura CZ" (e-INFRA LM2018140) provided within the program Projects of Large Research, Development and Innovations Infrastructures.

\bibliographystyle{acl_natbib}
\bibliography{anthology,ranlp2021}

% TODO tohle musi byt v samostanem filu
% TODO zkontrolovat odkud vsude se to referencuje a jestli se to referencuje spravne
\appendix
\newpage
\clearpage

% TODO ================
% ====================== POZOR APPENDIX MUSI BYT NAHRAN SAMOSTATNE ===========
\section{Appendix}
\subsection{LSTM Hyper-parameters}
\label{sec:apendix-hyper}
We use cross-entropy as the loss function and the Adam \cite{Kingma-adam} optimizer with default parameters ($\beta_1 = 0.9, \beta_2 = 0.999$) and with a modification from \cite{loshchilov2017decoupled} for the FB dataset. The embedding layer is trainable with a maximum size of 300k. The max sequence length for the input $t$ tokens is 64 for the FB dataset and 512 for the CSFD and Mallcz dataset with weight decay in the optimizer set to 0. We use Czech pre-trained fastText \cite{bojanowski-etal-2017-enriching-fasttext} embeddings\footnote{Available at \url{https://fasttext.cc/docs/en/crawl-vectors.html}}. For the Mallcz and CSFD datasets, we use 128 units in the BiLSTM layers and a batch size of 128. For the FB dataset, we use 256 units in the BiLSTM layers and a batch size of 256 with weight decay in the optimizer set to 1e-4.

For all datasets, we use 10 \% of total steps (batch updates) to warm up the learning rate, which means that during the training, the linear rate is firstly linearly increasing to the initial learning rate before being decayed with the corresponding learning rate scheduler. The dropout after the BiLSTM layers is set to 0.2. We use cosine (the * symbol in Table \ref{tab:transformer-hyperparameters}) and the exponential learning rate scheduler (the $\ddagger$ symbol in Table \ref{tab:transformer-hyperparameters}) with a decay rate set to 0.05. Table \ref{tab:transformer-hyperparameters} contains the initial learning rate and the number of epochs for the \texttt{LSTM} model for each dataset.

\subsection{Transformer Hyper-parameters}
\label{sec:transformer-appendix}
For fine-tuning of the transformer-based models, we use the same modification \cite{loshchilov2017decoupled} of the Adam \cite{Kingma-adam} optimizer with default weight decay set to 1e-2. We use different learning rates and a number of epochs for each combination of the models and datasets, see Table \ref{tab:transformer-hyperparameters}. We use either constant linear rate or linear learning rate decay without learning rate warm-up. We use default values of all other hyper-parameters.

\par For the cross-lingual experiments, we use only the linear learning rate decay scheduler with the initial learning rate set to 2e-6 without learning rate warm-up. For the cross-lingual experiments from English to Czech, the numbers of epochs used for the fine-tuning are 5, 2, 4 and 10 for the \texttt{XLM-R-Base, XLM-R-Large, XLM} and \texttt{mBERT} models, respectively. For the cross-lingual experiments from Czech to English, the numbers of epochs used for the fine-tuning are 25, 5 and 9 for the \texttt{XLM-R-Base, XLM-R-Large} and \texttt{XLM} models\footnote{The \texttt{mBERT} model did not converge for this experiment}, respectively.

\subsection{Computational Cluster}
For fine-tuning of the Transformers models we use the Czech national cluster Metacentrum\footnote{See \url{https://wiki.metacentrum.cz/wiki/Usage_rules/Acknowledgement}}. We use two NVIDIA A100 GPUs each with 40GB memory.

% https://pytorch.org/docs/stable/optim.html todo AdamW default weight decay
% https://stackoverflow.com/questions/64621585/pytorch-adamw-and-adam-with-weight-decay

\begin{table*}[ht!]
\catcode`\-=12
\begin{adjustbox}{width=\linewidth,center}
\begin{tabular}{llllllll}
\toprule
     \multirow{2}{*}{Model}            & \multicolumn{3}{c}{3 Classes}                                                                                            &       & \multicolumn{3}{c}{2 Classes}                                                                                                      \\ \cline{2-4} \cline{6-8}
                & \multicolumn{1}{c}{CSFD}                 & \multicolumn{1}{c}{FB}                   & \multicolumn{1}{c}{Mallcz}    &            & \multicolumn{1}{c}{CSFD}                   & \multicolumn{1}{c}{FB}                    & \multicolumn{1}{c}{Mallcz}                \\ \midrule
Log. reg. (ours)     & \multicolumn{1}{c}{79.63}               & \multicolumn{1}{c}{67.86}               & \multicolumn{1}{c}{76.71}         &       & \multicolumn{1}{c}{91.42}                 & \multicolumn{1}{c}{88.12}                & \multicolumn{1}{c}{88.98}                \\
LSTM (ours)           &     79.88 $\pm$ 0.18 \footnotesize{(5e-4 / 2)*}                                     & 72.89 $\pm$ 0.49  \footnotesize{(5e-4 / 5)*}                                       &          73.43 $\pm$ 0.12    \footnotesize{(5e-4 / 10) $\ddagger$}                    &           &   91.82 $\pm$ 0.09  \footnotesize{(5e-4 / 2)*}                                         &    90.13 $\pm$ 0.17        \footnotesize{(5e-4 / 5)*}                              &        88.02 $\pm$ 0.24    \footnotesize{(5e-4 / 2)$\ddagger$}                                  \\ \hdashline
Czert-A     & $79.89 \pm 0.60$  \footnotesize{(2e-6 / 8)}          & $73.06 \pm 0.59$  \footnotesize{(2e-5 / 8)}           & $76.79 \pm 0.38$  \footnotesize{(2e-5 / 12)}   &        & $91.84 \pm 0.84$  \footnotesize{(2e-5 / 8)}             & $91.28 \pm 0.18$  \footnotesize{(2e-5 / 15)}$\dagger$          & 91.20 $\pm$ 0.26  \footnotesize{(2e-5 / 14)}           \\
Czert-B       & 84.80 $\pm$ 0.10  \footnotesize{(2e-5 / 12)}          & 76.90 $\pm$ 0.38  \footnotesize{(2e-6 / 5)}$\dagger$          & 79.35 $\pm$ 0.24  \footnotesize{(2e-5 / 15)}       &    & 94.42 $\pm$ 0.15  \footnotesize{(2e-5 / 15)}            & 93.97 $\pm$ 0.30  \footnotesize{(2e-5 / 2)}            & 92.87 $\pm$ 0.15  \footnotesize{(2e-5 / 15)}           \\
mBERT           & 82.74 $\pm$ 0.16 \footnotesize{(2e-6 / 13)}           & 71.61 $\pm$ 0.13  \footnotesize{(2e-6 / 13)}$\dagger$         & 70.79 $\pm$ 5.74 \footnotesize{(2e-5 / 10)}   &         & 93.11 $\pm$ 0.29  \footnotesize{(2e-6  / 14)}$\dagger$          & 88.76 $\pm$ 0.42  \footnotesize{(2e-5 / 8)}            & 72.79 $\pm$ 3.09   \footnotesize{(2e-5 / 1)}           \\
SlavicBERT      & 82.59 $\pm$ 0.12  \footnotesize{(2e-6 / 12)}          & 73.93 $\pm$ 0.53  \footnotesize{(2e-5 / 4)}           & 75.34 $\pm$ 2.54  \footnotesize{(2e-5 / 10)}     &      & 93.47 $\pm$ 0.33  \footnotesize{(2e-6  / 15)}$\dagger$          & 89.84 $\pm$ 0.43  \footnotesize{(2e-5 / 9)}$\dagger$           & 90.99 $\pm$ 0.15  \footnotesize{(2e-6 / 14)}$\dagger$          \\
RandomALBERT   & 75.79 $\pm$ 0.18  \footnotesize{(2e-6 / 14)}           & 62.53 $\pm$ 0.46  \footnotesize{(2e-6 / 14)}$\dagger$         & 64.81 $\pm$ 0.25  \footnotesize{(2e-6 / 15)}$\dagger$     &     & 89.99 $\pm$ 0.21 \footnotesize{(2e-6  /  14)}$\dagger$           & 81.71 $\pm$ 0.56  \footnotesize{(2e-6 / 15)}$\dagger$          & 85.38 $\pm$ 0.10   \footnotesize{(2e-6 / 14)}$\dagger$         \\
XLM-R-Base      & 84.82 $\pm$ 0.10  \footnotesize{(2e-6 / 15)}$\dagger$           & 77.81 $\pm$ 0.50  \footnotesize{(2e-6 / 7)}$\dagger$          & 75.43 $\pm$ 0.07   \footnotesize{(2e-6 / 15)}$\dagger$      &   & 94.32 $\pm$ 0.34 \footnotesize{(2e-6  /14)} $\dagger$           & 93.26 $\pm$ 0.74 \footnotesize{(2e-6 / 5)}$\dagger$            & 92.56 $\pm$ 0.07  \footnotesize{(2e-6 / 12)}$\dagger$          \\
XLM-R-Large     & \textbf{87.08 $\pm$ 0.11}  \footnotesize{(2e-6 / 11 )} & \textbf{81.70 $\pm$ 0.64}  \footnotesize{(2e-6 / 5)}$\dagger$ & \textbf{79.81 $\pm$ 0.21}  \footnotesize{(2e-6 / 24)}$\dagger$ & & \textbf{96.00 $\pm$ 0.02} \footnotesize{(2e-6  / 143)}$\dagger$ & \textbf{96.05 $\pm$ 0.01}  \footnotesize{(2e-6 / 15)}  & \textbf{94.37 $\pm$ 0.02}  \footnotesize{(2e-6 / 15)}$\dagger$ \\
XLM             & 83.67 $\pm$ 0.11  \footnotesize{(2e-5 / 11)}           & 71.46 $\pm$ 1.58  \footnotesize{(2e-6 / 9)}$\dagger$          & 77.56 $\pm$ 0.08  \footnotesize{(2e-6 / 14)}$\dagger$       &   & 93.86 $\pm$ 0.18  \footnotesize{(2e-5 / 5)}             & 89.94 $\pm$ 0.27  \footnotesize{(2e-6 / 15)}$\dagger$          & 91.97 $\pm$ 0.22  \footnotesize{(2e-6 / 16)}$\dagger$          \\ \bottomrule
\end{tabular}
\end{adjustbox}
\caption{The final monolingual results as macro $ F_{1}$ score and hyper-parameters for all three Czech polarity datasets on two and three classes. For experiments with neural networks performed by us, we present the results with a 95\% confidence interval. For each result, we state the used learning rate and the number of epochs used for the training. The $\dagger$ symbol denotes that the result was obtained with constant learning rate, $*$ denotes the cosine learning rate decay, $\ddagger$ denotes exponential learning rate decay, otherwise the linear learning rate decay was used.} \label{tab:transformer-hyperparameters}
\end{table*}

\begin{table*}[ht!]
\begin{adjustbox}{width=\linewidth,center}
\begin{tabular}{llllll} \toprule
 \multirow{2}{*}{Model}             & \multicolumn{2}{c}{CSFD}              &                        & \multicolumn{2}{c}{FB}                                      \\ \cmidrule{2-3} \cmidrule{5-6}
             & \cite{sido2021czert}             & Ours       &                      & \cite{sido2021czert}                & Ours                             \\ \midrule
mBERT        & 82.80 ± 0.14 \footnotesize{(2e-6 / 13)} & 82.74 $\pm$ 0.16  \footnotesize{(2e-6 /  13)}  &    & 71.72 ± 0.91 \footnotesize{(2e-5 / 6)}  & 71.61 $\pm$ 0.13  \footnotesize{(2e-6 / 13)}  \\
SlavicBERT   & 82.51 ± 0.14 \footnotesize{(2e-6 / 12)} & 82.59 $\pm$ 0.12  \footnotesize{(2e-6 /  12)} &   & 73.87 ± 0.50 \footnotesize{(2e-5 / 3)}  & 73.93 $\pm$ 0.53  \footnotesize{(2e-5 / 4)}   \\
RandomALBERT & 75.40 ± 0.18 \footnotesize{(2e-6 / 13)}  & 75.79 $\pm$ 0.18  \footnotesize{(2e-6 / 14 )}  &    & 59.50 ± 0.47 \footnotesize{(2e-6 / 14)} & 62.53 $\pm$ 0.46  \footnotesize{(2e-6 / 14)}  \\
Czert-A      & 79.58 ± 0.46 \footnotesize{(2e-6 / 8)}  & 79.89 $\pm$ 0.60  \footnotesize{(2e-6 / 8)}  &    & 72.47 ± 0.72 \footnotesize{(2e-5 / 8)}  & 73.06 $\pm$ 0.59  \footnotesize{(2e-5 / 8)}   \\
Czert-B      & 84.79 ± 0.26 \footnotesize{(2e-5 / 12)} & 84.80 $\pm$ 0.10  \footnotesize{(2e-5 /  12 )} &   & 76.55 ± 0.14 \footnotesize{(2e-6 / 12)} & 76.90 $\pm$ 0.38  \footnotesize{(2e-6 / 5)}   \\ \bottomrule
\end{tabular}
\end{adjustbox}
\caption{Comparison of results from \cite{sido2021czert} with results obtained by us.} \label{tab:czert-res-comparison}
\end{table*}

\end{document}